\pdfoutput=1
\documentclass[10pt,twocolumn,letterpaper]{article}

\usepackage{3dv}
\usepackage{times}
\usepackage{epsfig}
\usepackage{graphicx}
\usepackage{amsmath}
\usepackage{amssymb}
\usepackage{amsthm}
\newtheorem{prop}{Proposition}

\newcommand{\diag}[1]{\mbox{diag}[#1]}
\renewcommand{\Re}{\mathbb{R}}

\usepackage[pagebackref=true,breaklinks=true,letterpaper=true,colorlinks,bookmarks=false]{hyperref}

\threedvfinalcopy
\begin{document}

\title{{\scriptsize \vspace{91em} To appear at the International Conference on 3D Vision (3DV), 2016.\hfill~\\\vspace{-92em}}%
Single-image RGB Photometric Stereo With Spatially-varying Albedo}
\author{Ayan Chakrabarti\\TTI-Chicago \and Kalyan Sunkavalli\\Adobe Research}

\maketitle
\thispagestyle{empty}

\begin{abstract}

We present a single-shot system to recover surface geometry of objects with spatially-varying albedos, from images captured under a calibrated RGB photometric stereo setup---with three light directions multiplexed across different color channels in the observed RGB image. Since the problem is  ill-posed point-wise, we assume that the albedo map can be modeled as piece-wise constant with a restricted number of distinct albedo values. We show that under ideal conditions, the shape of a non-degenerate local constant albedo surface patch can theoretically be recovered exactly. Moreover, we present a practical and efficient algorithm that uses this model to robustly recover shape from real images. Our method first reasons about shape locally in a dense set of patches in the observed image, producing shape distributions for every patch. These local distributions are then combined to produce a single consistent surface normal map. We demonstrate the efficacy of the approach through experiments on both synthetic renderings as well as real captured images.
\end{abstract}

\begin{figure*}[!t]
  \centering
  \includegraphics[width=0.95\textwidth]{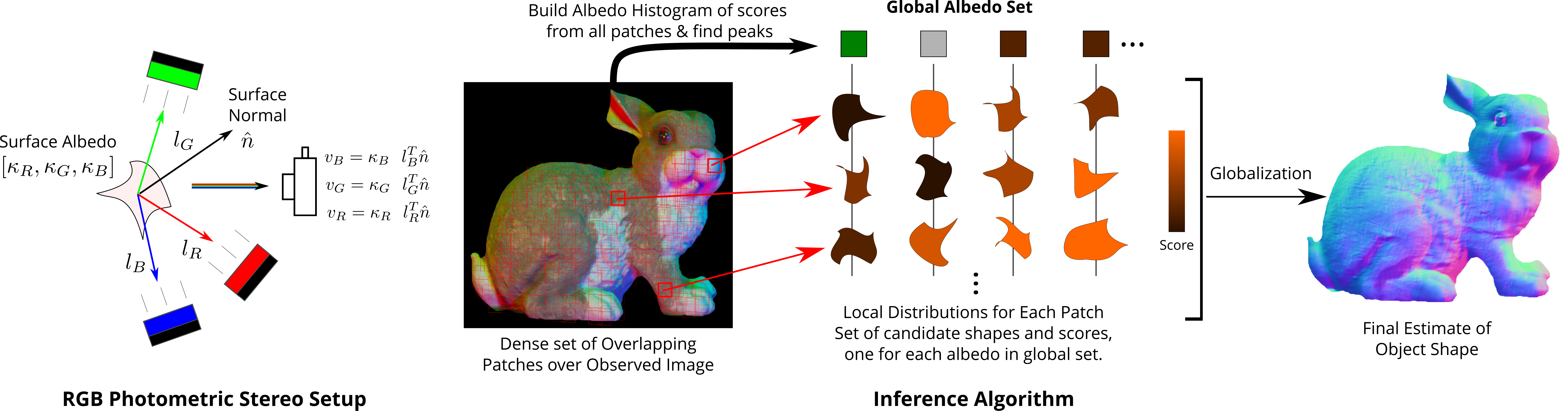}
  \caption{Overview of proposed system. (Left) RGB-PS capture: a diffuse surface is illuminated by three directional mono-chromatic light sources, with shading due to each light captured in a different color channel of an RGB camera. (Right) Inference using a piece-wise constant albedo model: we first perform local inference on a dense set of overlapping patches in the observed image, producing distributions of candidate shapes for every patch, where each candidate corresponds to a different assumed albedo from a global set of candidate albedos. These local distributions are then harmonized to produce a single consistent surface normal map for the object.}
  \label{fig:teaser}
\end{figure*}

\section{Introduction}
\label{sec:intro}

Photometric stereo (PS) techniques reconstruct surface geometry from shading, using images acquired under calibrated lighting. While other depth sensors---like those based on triangulation or time-of-flight---provide measurements of depth or distance, shading intensities are directly related to surface orientation. This makes PS the preferred choice for recovering high-resolution surface relief. However, classical PS requires capturing multiple images of an object under different illumination environments---a minimum of three images for Lambertian objects~\cite{Hayakawa94:PS,Woodham80:PS}---to be well constrained. Consequently, it is predominantly used for recovering shapes of static objects.

At the same time, reasoning about shape from shading (SFS) in a single image has been a classical problem in computer vision~\cite{Horn70:SFS}. While early work made restrictive assumptions---like known or constant albedo and known or directional lighting---Barron and Malik~\cite{Barron15:SIRFS} recently demonstrated that reasonable surface reconstructions are possible from a single image of an object with unknown spatially-varying albedo under unknown natural lighting. Although impressive given the inherent ambiguities in the SFS setup, their recovered geometries are typically coarse due to the use of strong smoothness priors, and their inference algorithm is computationally expensive. This is true even when known lighting is provided as input to their algorithm, primarily because it is designed to handle arbitrary and potentially ambiguous \emph{natural} illumination environments.

In this paper, we show that efficient and high-quality surface recovery from a single image is possible, when using a calibrated lighting environment that is specifically chosen to be directly informative about shape. Specifically, we use the RGB (or color) photometric stereo (RGB-PS) setup~\cite{Brostow11:VideoNormals,Kontsevich94:SFS,Woodham94:RGBPS}, where an object is illuminated by three monochromatic directional light sources, such that each of the red, green, and blue channels in the observed image is ``lit'' from a different direction. For natural lighting, directional diversity in color has been shown to be informative towards shape~\cite{Johnson11:SFS}. But the benefits of this lighting setup for shape recovery can be better understood by interpreting it as one that multiplexes the multiple images of classical PS into the different color channels of a single image. 

Strictly speaking, RGB-PS observations are as ambiguous as images under a single directional light. While each pixel now has observations from three directional lights, there are also now three unknown albedos---one for each channel. This is why previous methods using the RGB-PS setup have had to rely on assuming constant surface albedo~\cite{Brostow11:VideoNormals,Woodham94:RGBPS}, or on capturing additional information~\cite{Anderson11:Multi,Fyffe11:SingleShot}. In this work, we present a single-shot RGB-PS estimation method that can handle spatially varying albedo, by relying on spatial reasoning to resolve this ambiguity. 

We assume that the albedo of the observed surface is piece-wise constant and consists of a finite, but unknown, set of distinct albedo values. We motivate our method by showing that for a typical non-degenerate constant-albedo local surface patch, its shape is uniquely determined by ideal Lambertian observations in the RGB-PS setup. However, this still requires identifying which patches in the image have constant albedo and which straddle boundaries. Moreover, noise, shadows, and other non-idealities can render independent per-patch shape reconstructions unstable.  

Accordingly, we propose a robust algorithm based on the inference framework of \cite{qsfs}, which combines local patch-wise inference with a global harmonization step. We analyze a dense overlapping set of patches in the observed RGB-PS image, and extract local shape distributions for each after identifying a restricted set of possible albedo values for the object. We then combine these local distributions to recover a consistent estimate of global object shape. 

We show that the combination of a piece-wise constant albedo assumption and the observation model in our RGB-PS setup enable computationally efficient inference. Both the computation of local shape distributions and global object shape can be efficiently mapped to modern parallel architectures. We systematically evaluate this method on both synthetic and real images. We find that our approach, while also being significantly faster, yields higher quality reconstructions with much more surface detail than the approach of \cite{Barron15:SIRFS} designed for generic natural lighting. In fact, we show that our method approaches the accuracy of classical multi-image PS with the same set of lighting directions.

\section{Related Work}
\label{sec:rw} 

Formalized initially by Horn~\cite{Horn70:SFS}, the SFS problem has been the focus of considerable research over the last few decades~\cite{Zhang99:SFSSurvey,Durou08:SFSSurvey}. A remarkably successful solution to the problem was recently proposed by \cite{Barron15:SIRFS}, who introduced a versatile method to recover object shape from a single image of a diffuse object with spatially-varying albedo. However, since it was designed for general un-calibrated natural lighting, their inference algorithm is computationally expensive and relies heavily on strong geometric smoothness priors. In contrast, our method is designed for a  known optimized lighting setup, and is able to efficiently recover shape with a higher degree of surface detail.

RGB-PS was introduced as a means to overcome the requirement in classical PS of capturing multiple images, which makes the latter unusable on moving or deforming objects (although, some methods attempt to handle such cases using multi-view setups~\cite{Vlasic09:DynamicShape}). However, the degree of ambiguity (5 unknowns for 3 observations) in RGB-PS reconstruction~\cite{Kontsevich94:SFS,Woodham94:RGBPS} is the same as that in single image SFS (3 unknowns for 1 observation).  Previous work addressed this by disallowing albedo variations~\cite{Brostow11:VideoNormals,Johnson09:SFS}, or by exploiting the temporal constancy of surface reflectance~\cite{Zsolt10:Dynamic}. Anderson \etal~\cite{Anderson11:Multi} use a stereo rig with multiplexed color lights. They reconstruct coarse shape and align shading intensities using stereo. This is used to segment the scene into constant albedo regions, followed by albedo estimation and refinement of surface depth and orientation estimates.

An exception is the work of Fyffe \etal~\cite{Fyffe11:SingleShot}, who like us, rely on the statistics of natural albedos. They assume that surface albedo, as a function of spectral wavelength, is low-dimensional. Since this assumption doesn't provide an informative constraint for albedos in just three color channels, their setup involves multi-spectral capture under six spectrally distinct color sources. However, this requires a more complex imaging system and also suffers from lower light efficiency---since the visible spectrum now is split into six, instead of three, non-overlapping bands for both illumination and sensing. In contrast, we rely on the spatial, instead of spectral, statistics of albedos, and are able to employ regular three-channel RGB cameras.

Our estimation algorithm employs a similar computational framework as Xiong \etal~\cite{qsfs}, who used a combination of dense local estimation and globalization for traditional SFS, assuming known albedo and a single known directional light. Our goal is different---we seek to recover high resolution geometric detail in the presence of spatially-varying albedo, from images captured under the RGB-PS setup. To this end, we employ a piece-wise constant assumption on albedo which we show to be informative in our setup, while \cite{qsfs} assumed piece-wise smooth shape.

\section{RGB Photometric Stereo}
\label{sec:prelim}

The RGB-PS setup, illustrated in Fig.~\ref{fig:teaser} (left), uses color multiplexing to capture different lighting directions in a single image. An object is illuminated with three directional mono-chromatic light sources, where each light's spectrum is such that it is observed in only one of the three color channels (red, green, or blue) of the camera. We let $l_R,l_G,l_B\in\Re^3$ denote the product of the direction and scalar intensity of these lights, with directions chosen so that the lighting matrix $L=[l_R,~l_G,~l_B]\in GL(3)$ is invertible.

Assuming no noise, the observed RGB intensities $v(p)\in\Re^3$ of an un-shadowed Lambertian surface point are
\begin{equation}
  \label{eq:lamb}
  v(p)=[v_R(p),v_G(p),v_B(p)]^T
  =\diag{\kappa(p)}L^T\hat{n}(p),
\end{equation}
where $\hat{n}(p)\in\mathbb{S}^2$ the unit normal of the surface point imaged at image location $p=(x,y)$, and $\kappa(p)\in R^3$ is the corresponding RGB surface albedo vector. 

Both $\hat{n}(p)$ and $\kappa(p)$ are unknown, and can not be recovered point-wise from the  three observed intensities in $v(p)$ alone. Therefore, we further assume that the object has piecewise constant albedo, \ie, the image can be segmented into a set of regions $\{\Omega_1,\Omega_2,\ldots\}$ such that all points within each region have the same albedo: $\kappa(p) = \kappa_i, \forall p \in \Omega_i$.

This assumption is useful because, as we show next, if a region is correctly identified as having constant albedo, its shape and albedo are typically determined uniquely by the ideal diffuse intensity measurements in the RGB-PS setup.
\begin{prop}\label{prop:1}
Given noiseless observed intensities $v(p)$ at a set of locations $p\in\Omega$ on a diffuse surface patch known to have constant albedo, i.e., $\kappa(p) = \kappa_\Omega, \forall p\in\Omega$, the true surface normals $\{\hat{n}(p): p\in\Omega\}$ and common albedo $\kappa_\Omega$ are uniquely determined, if: 
{\setlength{\parskip}{-1em}%
\begin{enumerate}%
{\setlength{\itemsep}{0pt}\setlength{\parskip}{0pt}%
\item All intensities $v(p)$ are strictly positive.
\item The true surface is non-degenerate in the sense that the set $\{\hat{n}(p)\hat{n}(p)^T: p\in\Omega\}$, of outer-products of the true normal vectors, span the space $\mbox{Sym}_3$ of all $3\times 3$ symmetric matrices.
}\end{enumerate}}\vspace{-1em}
\end{prop}

\noindent {\bf Proof}:~Given $\kappa_\Omega$ and $\hat{n}(p)$ as the true patch albedo and normals, let $\kappa'_\Omega$,~$\hat{n}'(p)$ be a second solution pair that also explains the observed intensities $v(p)$ in the patch $\Omega$. Since the observed intensities are strictly positive, this implies that the albedos $\kappa_\Omega,\kappa'_\Omega$ are strictly positive as well, and further that no point is in shadow under any of the lights, \ie~$L^T\hat{n}(p),L^T\hat{n}'(p) > 0,~\forall p \in \Omega$. Then, since $L^T$ is invertible, we can write
\begin{eqnarray}
  \label{eq:pf1}
  \diag{\kappa_\Omega}L^T\hat{n}(p) = \diag{\kappa'_\Omega}L^T\hat{n}'(p)\notag\\
  \Rightarrow \hat{n}'(p) = A\hat{n}(p),~~\forall p\in\Omega,
\end{eqnarray}
where we define the matrix $A=L^{-T}RL^T$, with $R=\diag{\kappa'_\Omega}^{-1}\diag{\kappa_\Omega}$ being a diagonal matrix whose entries are the ratio between the two albedo solutions. Note that these entries also correspond to the eigenvalues of $A$, and are real and positive since $\kappa_\Omega,~\kappa'_\Omega$ are real and positive.

Since $\hat{n}'(p)$ are unit vectors, we have as conditions on $A$:
\begin{eqnarray}
  \label{eq:lpres}
\|\hat{n}'(p)\|^2 = \|A\hat{n}(p)\|^2 = 1~\Rightarrow \hat{n}(p)^T~(A^TA)~\hat{n}(p) = 1\hspace{-2.5em}\notag\\
\Rightarrow\sum_{i,j} \left[(\hat{n}(p)\hat{n}(p)^T)\circ(A^TA) \right]_{ij} = 1,~\forall p\in\Omega,
\end{eqnarray}
where $\circ$ refers to the element-wise Hadamard product, and $[Q]_{ij}$ to the $(i,j)^{th}$ element of the matrix $Q$. \eqref{eq:lpres} represents a set of linear equations on the elements of $A^TA$. Since $A^TA \in \mbox{Sym}_3$ and $\{\hat{n}(p)\hat{n}(p)^T\}$ spans $\mbox{Sym}_3$, this linear system is full rank, and can have at most one solution. It is easy to see that this unique solution is given by $A^TA=I$.

Therefore, $A$ must be an orthogonal matrix, and since the only orthogonal matrix with only real and positive eigenvalues is the identity, $A=I$. This in-turn implies $R=I$, $\kappa'_\Omega=\kappa_\Omega$, and from \eqref{eq:pf1}, $\hat{n}'(p)=\hat{n}(p),~\forall p\in\Omega$. Therefore, the solutions must be identical, and the albedo and normals of the surface are uniquely determined.\hfill$\blacksquare$\\

Intuitively, the non-degeneracy condition in Prop.~\ref{prop:1} requires that the region $\Omega$ have sufficient diversity in its surface normals---such that no non-trivial linear transform preserves the length of all normal vectors. The curvature and relief in most surface regions typically render them non-degenerate. An example of a degenerate surface is a perfect plane---all normals, and therefore all observed intensities, in a plane are identical, and its orientation has the same ambiguity as the normal of a single point.

\section{Shape Estimation}
\label{sec:method}

It is important to remember that the uniqueness result in the previous section holds only in the ideal case. With observation noise, for example, there may be multiple diverse surface-albedo explanations that come equally close to explaining the intensities in a region. Moreover, a segmentation of the image into constant-albedo regions is not available, and must be inferred jointly estimated with the albedo and shape of each region. In this section, we describe a robust and efficient algorithm to carry out such inference. Formally, we seek to recover the surface normal field $\hat{n}(p)$ of an object, given its color image $v(p)$ acquired under known lighting $L$. Our estimation method uses a framework similar to that of \cite{qsfs}, with local inference followed by globalization to simultaneously reasons about whether different local patches have constant albedos, and if so, about their shape using the constant-albedo constraint.

Broadly, we consider a set of dense overlapping fixed-size patches in the observed image, and run local inference independently, and in-parallel, on each patch. For robustness, rather than commit to a single local shape estimate, the local inference step produces a distributional output for every patch in the form of a discrete set of candidate shapes and associated scores. Each shape in this distribution corresponds to surface normals that best explain the observed patch intensities assuming a different constant patch albedo. This set is computed for a dense sampling of the albedo space. Following Prop.~\ref{prop:1}, patches with good surface variation produce more accurate shape estimates and have scores that are tightly clustered around the correct albedo value. We further assume that the observed object overall has a limited number of distinct albedos. This adds an additional constraint to inference that makes the per-patch local distributions compact. We identify a global albedo set as the peaks of a histogram of scores over all albedos, computed from all patches in the observed image. Local distributions for each patch are then restricted to consist of shape estimates corresponding only to albedos in this global set. This allows albedo estimates in patches that are not ambiguous to restrict the space of solutions at other patches that are.

Local inference is followed by global optimization that finds a consistent normal map for the object by harmonizing the local shape distributions of all overlapping patches. This optimization refines the patch-wise shape estimates by considering consistency with other overlapping patches, and either selecting one of the local candidate shapes or deciding to ignore the local estimates all together to account for the possibility that the patch albedo is \emph{not} constant. We next describe the local inference and globalization steps in detail.

\subsection{Local Inference}

Given a dense set of patches $\{\Omega_m\}_{m=1}^M$ that cover the image plane, local inference produces distributions comprising sets of $K$ surface normal estimates $\{\hat{n}_{m:k}(p)\}_{k=1}^K,~\forall p\in\Omega_m$, and corresponding scores $\{s_{m:k}\}_{k=1}^K$, for every patch. These distributions are computed with respect to a global albedo set $\{\kappa_k\}_{k=1}^K$ for the image, where normal estimates $\hat{n}_{m:k}(p)$ are computed assuming $\kappa_k$ as the albedo in patch $\Omega_m$.

\paragraph{Local Shape Model}During inference, we represent each set of local surface normals $\hat{n}_{m:k}(p)$ based on a polynomial model for surface depth within a patch:
\begin{equation}
  \label{eq:zfromA}
z_{m:k}(p) = \underset{d_x,d_y\geq 0,1\leq d_x+d_y \leq D}{\sum} (a_{m:k})_{[d_x,d_y]}~x^{d_x}y^{d_y},
\end{equation}
where $D$ is the polynomial degree. The coefficient vector $a_{m:k}=[\{(a_{m:k})_{[d_x,d_y]}\}]^T$ describes the $k^{th}$ candidate shape estimate for $\Omega_m$, with $\hat{n}_{m:k}(p)$ corresponding to the surface normals of $z_{m:k}(p)$ above. This approach  automatically constrains each candidate normal set $\hat{n}_{m:k}(p)$ to be integrable. Note that a similar polynomial model (with $D=2$) was also used in \cite{qsfs}. However, unlike \cite{qsfs}, our goal is not to impose smoothness on our local shape estimates, but to make shape estimation more efficient. Therefore, we employ higher degree polynomials to able to express  high-frequency local relief.

We use $\tilde{n} \in \Re^2$ to represent the co-ordinates of the intersection of a normal vector $\hat{n}$ with the $z=1$ plane, \ie $\hat{n}=[\tilde{n},1]^T/\|[\tilde{n},1]\|$. These co-ordinates correspond to the gradients of depth: $\tilde{n} = [\partial z/\partial x, \partial z/\partial y]$. We let $n_{m:k}\in\Re^{2\|\Omega_m\|}$ denote a vector formed by concatenating the gradient vectors $\tilde{n}_{m:k}(p)$ at all pixels $p\in\Omega_m$. Then, assuming all patches $\Omega_m$ are the same size and using a patch-centered co-ordinate system in \eqref{eq:zfromA}, we have
\begin{align}
  \label{eq:nga}
  n_{m:k} &= \left[\ldots,~\tilde{n}_{m:k}(p_i),~\cdots\right]^T\notag\\
  &= \left[\begin{array}{ccc}&\vdots\\
      \cdots & d_xx_i^{d_x-1}y_i^{d_y}&\cdots\\
      \cdots&d_yx_i^{d_x}y_i^{d_y-1}&\cdots\\
&\vdots\end{array}\right]\left[\begin{array}{c}\vdots\\(a_{m:k})_{[d_x,d_y]}\\\vdots\end{array}\right]\notag\\
  &= Ga_{m:k},
\end{align}
\ie, the concatenated gradient vectors $n_{m:k}$ for all patches are related to their coefficients $a_{m:k}$ by the same matrix $G$.

\paragraph{Albedo Parameterization}For the albedo of each patch, we search over a discrete set formed by quantizing the space of possible albedo vectors $\kappa$. In particular, we factor $\kappa=\tau\hat{\kappa}$ as the product of a scalar ``luminance'' $\tau$, and a chromaticity vector $\hat{\kappa}$---the latter constrained to be a unit vector with non-negative elements. This factorization will prove convenient since the point-wise ambiguity in RGB-PS is resolved when the albedo chromaticity $\hat{\kappa}$ is known. We construct our discrete candidate albedo set by quantizing $\tau$ and $\hat{\kappa}$ separately into uniformly spaced bins $\{\tau_l\}_{l=1}^L$ in $[0,\tau_{\max}]$ and $\{\hat{\kappa}_c\}_{c=1}^C$ in $\mathbb{S}^2_+$ respectively. Here, the value of $\tau_{\max}$ depends on the scale of the intensities $v(p)$ and lighting matrix $L$. 

\paragraph{Global Albedo Set}The first step in inference is identifying a restricted global set of possible albedo values present in the object by pooling evidence from all patches.  We do this by iterating over the discrete values of candidate albedo chromaticities $\{\hat{\kappa}_c\}$, and for each chromaticity $\hat{\kappa}_c$, computing estimates of the albedo luminance values $\tau_{m:c}$ and integrable surface normals $\hat{n}_{m:c}(p), p \in \Omega_m$, for every patch $\Omega_m$. We score these estimates in terms of a normalized rendering error $s_{mc} = \mathcal{S}_m(\hat{\kappa}_c\times\tau_{m:c},\hat{n}_{m:c})$ where
\begin{equation}
  \label{eq:score}
  \mathcal{S}_m(\kappa,\hat{n}) = \frac{\sum_{p\in\Omega_m}\|v(p)-\diag{\kappa}L^T\hat{n}(p)\|^2}{\sum_{p\in\Omega_m}\|v(p)\|^2}.
\end{equation}
Note that the denominator above is the same for different albedo-shape explanations for a given patch, and only serves to weight contributions from different patches.

While we could use a full non-linear optimizer to minimize \eqref{eq:score} to  compute the luminance and normal estimates $\tau_{m:c}$ and $\hat{n}_{m:c}(p)$, we find that a much simpler and faster approach suffices. We first compute normals $\hat{n}^0_{c}(p)$ and luminance values $\tau_{c}(p)$ for individual pixels simply as
\begin{equation}
  \label{eq:taun}
  \tau_{c}(p)\times\hat{n}^0_{c}(p) = L^{-T}\diag{\hat{\kappa}_c}^{-1}v(p),
\end{equation}
disambiguating the two terms in the LHS using the fact that $\tau_{c}(p)$ is a scalar, and $\hat{n}^0_{c}(p)$ a unit vector. This computation can be done for all pixels efficiently on modern parallel architectures, since it maps to the product of the same matrix $(L^{-1}\diag{\hat{\kappa}_c}^{-1})$ with all the intensity vectors $v(p)$.

Then, we compute per-patch luminance and normal estimates by ``projecting'' their set of pixel-wise values to the constant-albedo and polynomial depth models respectively. We set $\tau_{m:c}$ simply to the mean of the corresponding pixel luminances $\{\tau_c(p),~p\in\Omega_m\}$ in the patch. For the normals, we find the best fit of the pixel-wise normals to the polynomial model for each patch 
\begin{equation}
  \label{eq:nproj}
a_{m:c} = (G^TG)^{-1}G^T~n^0_{m:c},
\end{equation}
where $n^0_{m:c}$ is the concatenated gradient vector for patch $\Omega_m$ formed from the per-pixel normals $\hat{n}^0_c(p), p\in\Omega_m$. We then set $\hat{n}_{m:c}(p)$ to the unit normals corresponding to  $n_{m:c}=Ga_{m:c}$. These computations can also be carried out efficiently, in this case parallelized across patches.

Using these estimates and corresponding errors $s_{m:c}$, we construct a global histogram $H[l,c]$ over the full discrete candidate albedo set using clipped values of these errors as
\begin{equation}
  \label{eq:hdef}
  H[l,c] = \sum_{m} I[\tau_{m:c} =_q \tau_l]\times\max\left(0,h_{\max}-s_{m:c}\right),
\end{equation}
where $I_q[\tau_{m:c} =_q \tau_l]$ is one when the quantized value of $\tau_{m:c}$ equals $\tau_l$, and zero otherwise. Every patch thus makes a contribution to only one luminance bin for every chromaticity value.  $H[l,c]$ represents a soft aggregation of the number of patches that have low rendering errors (as per  $h_{\max}$) for each albedo. Using non-maxima suppression, we construct our global albedo set $\{\kappa_k = \tau_k\times\hat{\kappa}_k\}_{k=1}^K$ as the $K$ highest-valued peaks in the histogram $H[l,c]$.

\paragraph{Local Shape Distributions} We then recompute normal estimates and rendering error scores for all patches, now with respect to only the global albedo set. We follow a similar procedure as above. We iterate over the chromaticities of the albedos in the global set, and for each $\hat{\kappa}_k$, we compute pixel-wise luminance and normal values $\tau_k(p)$ and $\hat{n}^0_k(p)$ using \eqref{eq:taun}. We compute the per-patch surface coefficients $a_{m:k}$, and therefore the corresponding normals $\hat{n}_{m:k}(p)$, from $\hat{n}^0_k(p)$ using \eqref{eq:nproj}. Our local distributions are then $K$ pairs $\{a_{m:k},s_{m:k}\}_{k=1}^K$ of these surface coefficients, along with rendering scores $s_{m:k}=\mathcal{S}_m(\hat{\kappa}_k\times \tau_{m:k},\hat{n}_{m:k})$. Here, we set $\tau_{m:k}$ by projecting the mean, of the per-pixel luminances $\{\tau_k(p),~p\in\Omega_m\}$, to the bin corresponding to the luminance $\tau_k$ of the $k^{th}$ albedo in the global set.

\begin{figure*}[!t]
  \centering
  \includegraphics[width=0.74\textwidth]{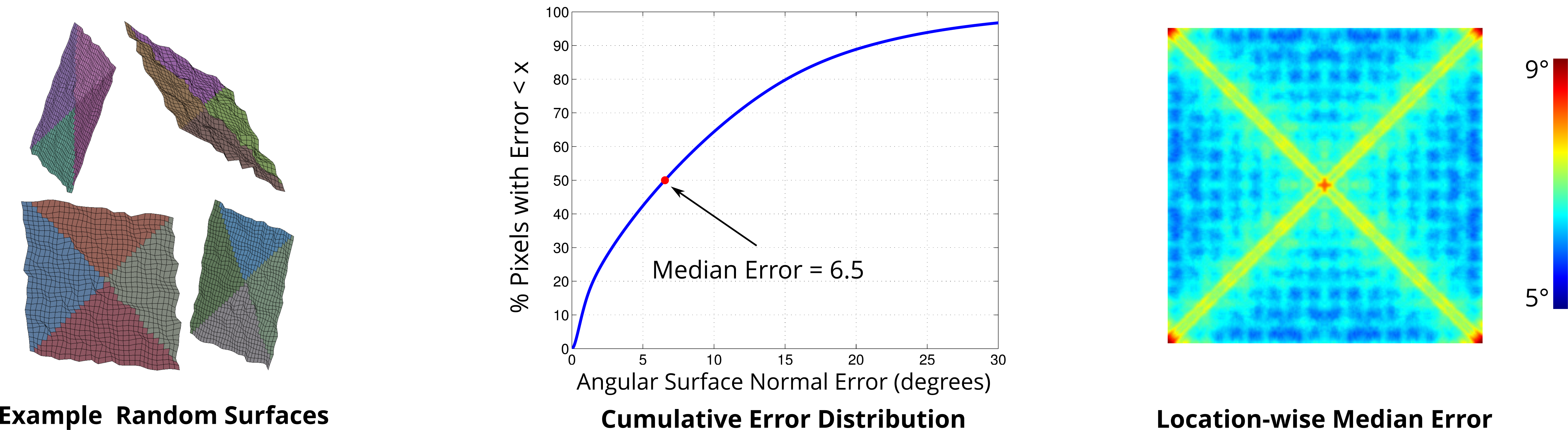}
  \caption{Quantitative Evaluation on Synthetic Surfaces. We evaluate our method on synthetically rendered images of a thousand randomly generates surfaces (left). We show overall statistics of estimation error (center), as well as the spatial distribution of these errors, which indicates that errors are only slightly higher near albedo boundaries (right).}
  \label{fig:synthFig}
\end{figure*}

\subsection{Global Shape Estimation}

To form our final shape estimate, we have to find a single shape estimate for each patch $\Omega_m$---by deciding between selecting one of multiple shape candidates, or ignoring them all together to account for patches with varying albedo---and harmonize normal estimates at each pixel $p$ from multiple overlapping patches that include it. 

We do this by employing an alternating iterative algorithm to minimize a consensus-based cost function similar to \cite{qsfs}. This cost function is defined over the pixel-wise depth gradient map $n(p)$, and auxiliary variables $\{a_m\}$ that correspond to per-patch shape coefficients,  as
\begin{align}
  \label{eq:conobj}
  &\mathcal{L}(n(p),\{a_m\}) =\sum_{m=1}^M\Bigg[ \lambda~\|n_m-Ga_m\|^2\notag\\
  &+\min\left(\gamma,\min_k\left( s_{m:k} + \|G(a_{m} - a_{m:k})\|^2\right)\right)\Bigg],
\end{align}
where $n_m$ is formed from concatenating $n(p), p \in \Omega_m$. The first term of the cost function essentially requires the gradients $n(p)$ at each pixel to be close to their predicted estimates $Ga_m$ from all patches $\Omega_m \ni p$ that include that pixel. The second term enforces fidelity between the per-patch shape coefficients $a_m$ and the local distributions $\{(s_{m:k},a_{m:k})\}$. $\lambda$ is a scalar parameter that controls the relative contribution of these two terms.

The fidelity of $a_m$ to each candidate shape $a_{m:k}$ is defined as the sum of the squared error between them and the shape's score $s_{m:k}$. \eqref{eq:conobj} considers the best cost across the different candidates for each patch, and to be able to reject distributions for patches with varying albedo, applies a threshold $\gamma$. When costs of all candidates are beyond this threshold, $a_m$ no longer depends on any of the candidate shapes $\{a_{m:k}\}$. Note ``outlier'' handling in our setup serves a different purpose  than it did in \cite{qsfs}. While we ignore the shape candidates for an outlier, we still enforce the local polynomial shape model. Thus, we only reject the constant-albedo assumption. Our higher-degree polynomial shape model encodes integrability, not smoothness like in \cite{qsfs}, and enforcing it even in outlier patches allows us to avoid a separate global integrability term in the objective in \eqref{eq:conobj}.

We minimize \eqref{eq:conobj} using an iterative algorithm that alternates between optimizing with respect to $n(p)$ and to $\{a_m\}$, while keeping the other fixed. We also find it useful to begin the iterations with a smaller value of $\lambda$, and increase it by a constant factor at each iteration till it reaches its final value. We begin by initializing each $a_m$ to simply the candidate shape $a_{m:k}$ with the smallest value of $s_{m:k}$. Then, in each iteration, we first minimize with respect to the gradient map $n(p)$ keeping $\{a_m\}$ fixed. This is achieved simply by setting each $n(p)$ to the mean of its estimates $\{(Ga_{m})(p)\}_{m:\Omega\ni p}$ from all patches containing $p$. 

The second step at each iteration minimizes \eqref{eq:conobj} with respect to $\{a_m\}$, which can be done independently for each $a_m$. We first compute a set of auxiliary coefficients and scores based on $n(p)$ as $\bar{a}_{m:0} = (G^TG)^{-1}G^Tn_m,~~\bar{s}_{m:0} = \gamma,$ and for $k\in\{1,\ldots K\}$, $\bar{a}_{m:k}= (1+\lambda)^{-1}(a_{m:k}+\lambda a_{m:0}),~\bar{s}_{m:k}= s_{m:k} + \|G(\bar{a}_{m:k}-a_{m:k})\|^2$. Each $a_m$ is then set to the $\bar{a}_{m:k}$ among $k\in\{0,\ldots K\}$ (\ie, including the outlier case $\bar{a}_{m:0}$) for which $\bar{s}_{m:k}$ is lowest.

\section{Experimental Results}
\label{sec:exp}

We now report quantitative and qualitative results on the performance of the proposed method on a large number of synthetically generated surfaces, as well as on acquired images of real objects. In all experiments, we use fully overlapping sets of $8\times 8$ patches. For the polynomial shape model, we choose degree $D=5$, and for albedo discretization, we choose 4096 bins for chromaticity---$64$ each over elevation and azimuth of $S^2_+$---and 100 bins for luminance, and set $\tau_{\max}=3$ for observed intensities in the range $[0,1]$. 

For local inference, we consider a global albedo set of size $K=100$, and set the histogram error threshold $h_{\max}$ to $10^{-4}$ for the synthetic surface renderings in Sec~\ref{sec:synth} below, and to a higher value of $10^{-2}$ for the real acquired images in Sec.~\ref{sec:real} to account for higher noise and other non-idealities. For global inference, we set the outlier threshold $\gamma=4$. We run alternating minimization starting with $\lambda=2^{-64}$, increasing it by a constant factor of $\sqrt{2}$ at each iteration till it reaches $256$, for a total of 145 iterations.

\subsection{Synthetic Images}
\label{sec:synth}

We synthetically render $1000$ randomly generated surfaces to conduct a systematic quantitative evaluation of our method's performance. Each image is of size $256\times 256$ pixels, and is rendered using randomly generated albedo and depth maps and a common chosen lighting $L$. The albedo map is generated by dividing the image into four equal triangles, and picking a random albedo vector per triangle. The surface is generated by first choosing a random base planar (ensuring that it is not in shadow), and adding zero-mean Gaussian depth perturbations---generated first at a coarser scale (of $16\times 16$) and smoothly up-sampled to $256\times 256$. Examples of these random surfaces are in Fig.~\ref{fig:synthFig} (left).

We render all surfaces using \eqref{eq:lamb} with moderate $0.1\%$ Gaussian observation noise, simulating attached shadows by clipping negative values of $L^T\hat{n}$ to zero. We run our full algorithm on each image, and compute angular errors between estimated and true surface normals. Figure~\ref{fig:synthFig} (center) shows a cumulative distribution of these errors across all pixels in all surfaces---summarizing our estimation accuracy over a diverse set of albedo-geometry combinations. We see that our method is usually able to recover accurate surface geometry, with a median error of $6.5^\circ$. As the albedo boundaries in all our rendered images are aligned, we are also able to visualize how performance varies in pixels close to these boundaries. Figure~\ref{fig:synthFig} (right) shows location-wise median errors, \ie median across surfaces of errors at each pixel location. As expected, we see that errors are higher near albedo boundaries. However, the range of this variation is small---from roughly $5^\circ$ within constant albedo regions to a high of $9^\circ$ at albedo ``corners''.

\subsection{Real Images}
\label{sec:real}

\begin{figure*}[!t]\centering
  \includegraphics[width=0.87\textwidth]{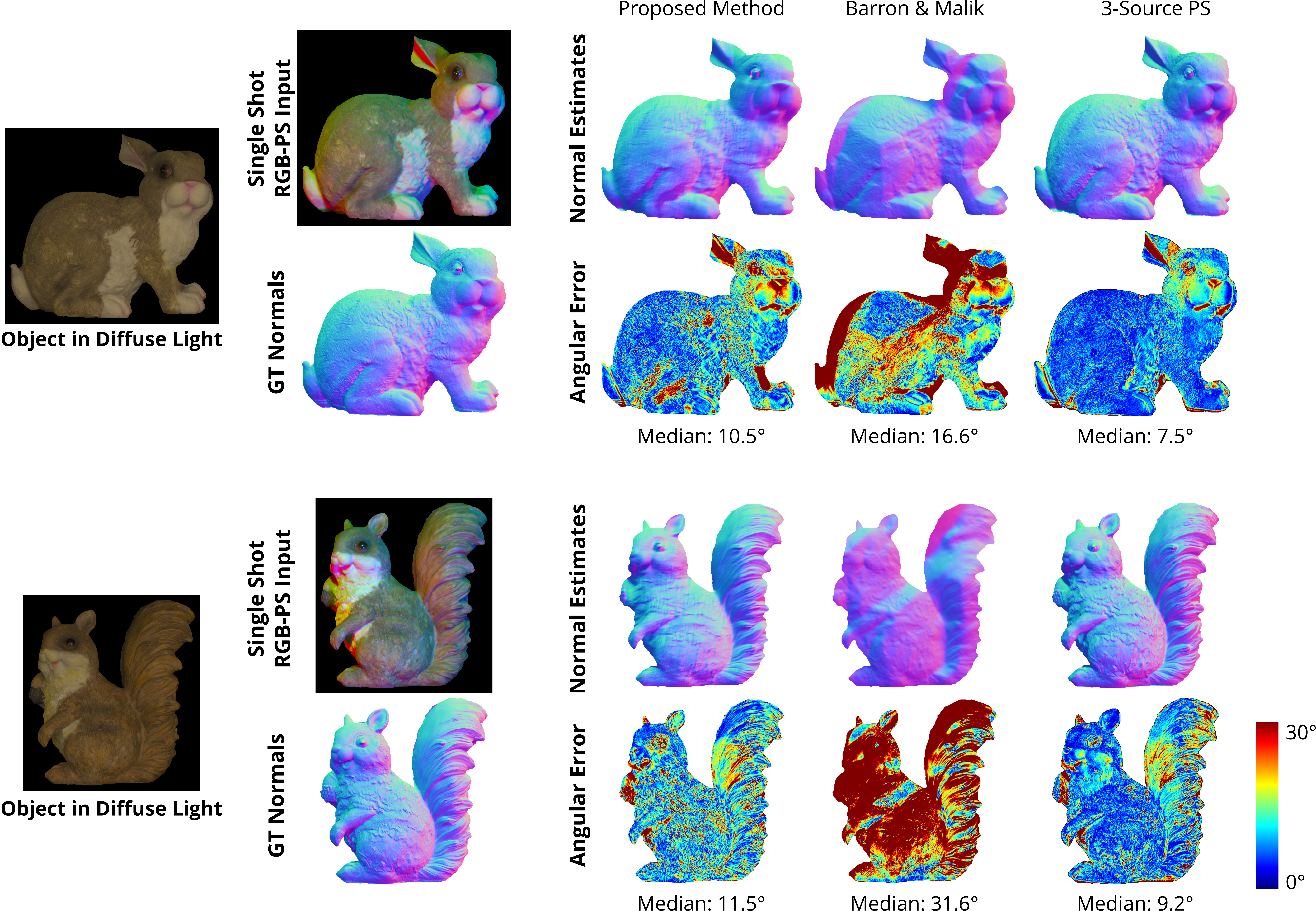}
  \caption{Results on single-shot captured images of real objects. We show estimated normal maps, and corresponding errors, for our method as well as that of Barron \& Malik~\cite{Barron15:SIRFS}. As comparison, we also show results for running classical photometric stereo on three full-color images captured under the same lighting directions $L$ (simulated using known ground-truth albedo). Errors in these estimates are due to shadowing and other non-idealities, and thus they provide an upper-bound to our performance.}
  \label{fig:realFig}
\end{figure*}

\begin{figure*}[!t]\centering
  \includegraphics[width=0.87\textwidth]{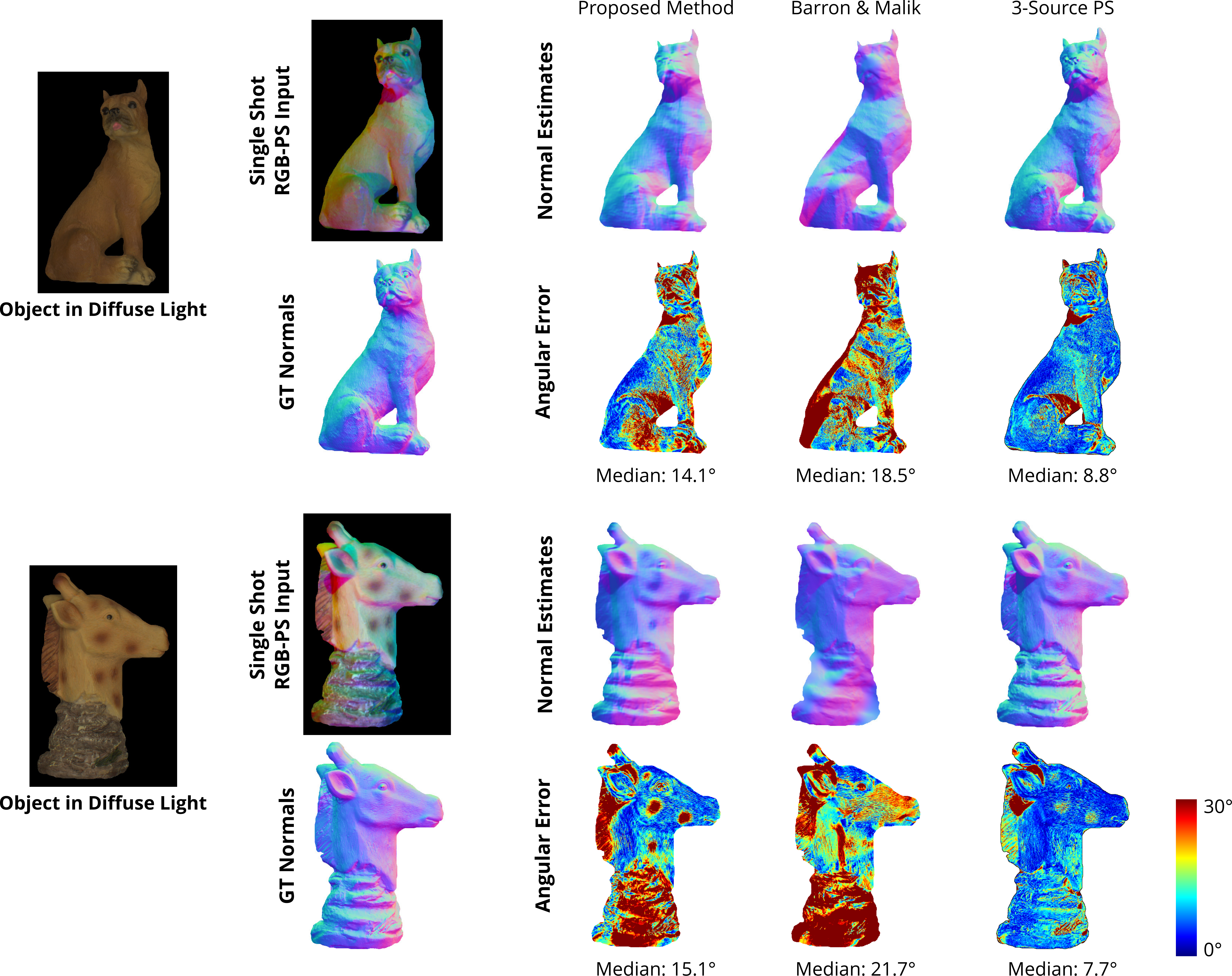}
  \caption{Results on real objects (continued).}
  \label{fig:realFig2}
\end{figure*}

\begin{figure*}[!t]
  \centering
  \includegraphics[height=0.17\textwidth]{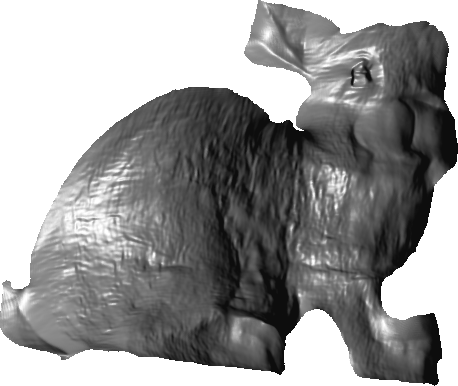}
  \hspace{0.05\textwidth}
  \includegraphics[height=0.17\textwidth]{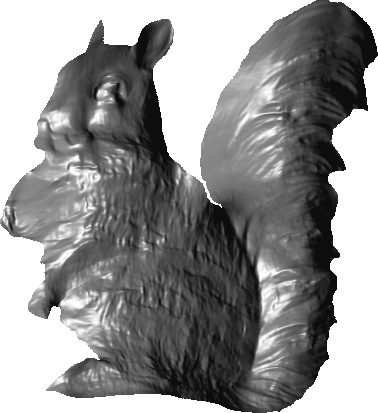}
  \hspace{0.05\textwidth}
  \includegraphics[height=0.17\textwidth]{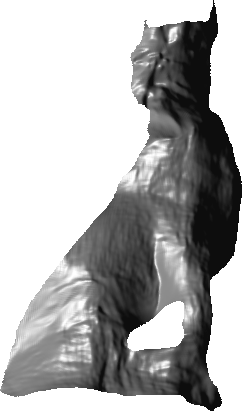}
  \hspace{0.05\textwidth}
  \includegraphics[height=0.17\textwidth]{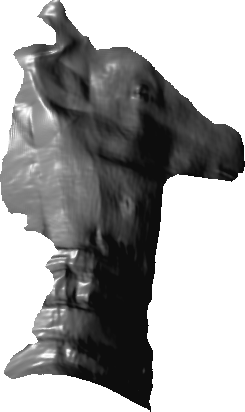}
  \caption{Simulated alternate views using integrated  depth maps from our normal estimates.}
  \label{fig:realFig3}
\end{figure*}

We evaluate our method on four real objects that were single-shot imaged by a Canon EOS 40D camera under our RGB photometric stereo setup. We place color filters in front of three LED lights, with filters chosen to create monochromatic lights---we ensure that in a scene lit by, say, only the red light source, green and blue camera intensities are nearly zero. Lighting directions $L$ are estimated with a chrome sphere. We work with RAW camera images, where color channels are multiplexed using a Bayer pattern. To avoid artifacts from demosaicking, we blur the image with a one pixel std.~Gaussian filter for anti-aliasing, and then down-sample to form a single RGB pixel for every $2\times 2$ Bayer block. We compute an object mask against the dark background by simple thresholding, and only run inference within this mask. We also white-balance each image (dividing each channel by its mean intensity), so that the discretization of our albedo search space is uniform.  

We also capture images of each object from the same camera, now under different directional white light sources, and run robust classical photometric stereo to get aligned ground truth normal and albedo maps. Moreover, for comparison, we use the known albedo and single-shot RGB image to simulate three separate captures under white lights with the same exact directions as our setup. We estimate a set of surface normals through classical photometric stereo on these images. Since errors in these normals are due only to non-idealities like shadowing, inter-reflections, specularities, \etc, they represent an upper bound on the performance under our more ambiguous single-shot setup.

Figures~\ref{fig:realFig} and \ref{fig:realFig2} show our results on these real objects, and Fig.~\ref{fig:realFig3} shows alternate views rendered using depth maps obtained by integrating our estimated normals. Our method produces high-quality surface normal estimates in most regions, even though the objects feature natural albedo variations that deviate from our strict piecewise constant model. This highlights the robustness of our method, and its practical utility. Indeed, we find that most of our errors are in regions where three-source photometric stereo also fails (\eg, due to shadows), although these errors are exaggerated in our estimates---both in magnitude and spatial extent. Also note the errors in the base and dark spots of the ``giraffe'' in Fig.~\ref{fig:realFig2}. The albedo values in these regions have roughly constant chromaticity, but continuously changing luminance---and happen to provide a plausible, but incorrect, solution under the piecewise constant albedo model.

We also include results from \cite{Barron15:SIRFS} in Figs.~\ref{fig:realFig} and \ref{fig:realFig2}, providing it our calibrated lighting environment, but without using contour information.  We see that \cite{Barron15:SIRFS} recovers only a coarse estimate of surface geometry, with much less detail than our method. Moreover, it takes 20 mins.~for a $800\times 730$ image with $55\%$ valid pixels on a 6-core 3.5GHz CPU. In contrast, our method only takes 160 secs.~with a Titan X GPU. 

The source code for our implementation, along with data, is available for download at the project website at \url{http://www.ttic.edu/chakrabarti/rgbps/}.

\section{Conclusion}
\label{sec:conc}

In this paper, we presented a single-shot system for recovering the shape of objects with spatially-varying albedo, using a calibrated RGB-PS setup for acquisition. Inference was based on a piece-wise constant model for surface  albedo. We characterized the shape information in RGB-PS observations under this model, showing that exact recovery is possible under idealized conditions. Then, we described a robust and efficient inference algorithm that achieved high-quality results on complex real-world objects.

Our system's ability to perform accurate single-shot shape recovery means that it can be used to reconstruct dynamic, deforming objects from a sequence of video frames--which previously had required multi-view setups~\cite{Vlasic09:DynamicShape}. Beyond simply generating stand-alone shape estimates from each image, in future work we will explore efficient ways to incorporate temporal constraints across frames. We believe this can allow high-quality time-varying reconstructions from monocular video, for example, by ameliorating the effects of shadows---regions that are in shadow in some frames may be lit in others. We are also interested in extending our method to leverage additional information, like contours, when available, and incorporating non-Lambertian reflection models for complex materials.

\flushleft\textbf{Acknowledgments} AC was supported by NSF award no. IIS-1618021, a gift from Adobe, and a hardware donation from NVIDIA Corporation. KS thanks Fabian Langguth and Sunil Hadap for helpful discussions.

{\small

\begin{thebibliography}{10}\itemsep=-1pt

\bibitem{Anderson11:Multi}
R.~Anderson, B.~Stenger, and R.~Cipolla.
\newblock Color photometric stereo for multicolored surfaces.
\newblock In {\em Proc.~ICCV}, 2011.

\bibitem{Barron15:SIRFS}
J.~T. Barron and J.~Malik.
\newblock Shape, illumination, and reflectance from shading.
\newblock {\em IEEE Trans. PAMI}, 2015.

\bibitem{Brostow11:VideoNormals}
G.~J. Brostow, C.~Hern{\'a}ndez, G.~Vogiatzis, B.~Stenger, and R.~Cipolla.
\newblock Video normals from colored lights.
\newblock {\em IEEE Trans. PAMI}, 2011.

\bibitem{Durou08:SFSSurvey}
J.-D. Durou, M.~Falcone, and M.~Sagona.
\newblock Numerical methods for shape-from-shading: A new survey with
  benchmarks.
\newblock {\em Comput. Vis. Image Underst.}, 2008.

\bibitem{Fyffe11:SingleShot}
G.~Fyffe, X.~Yu, and P.~Debevec.
\newblock Single-shot photometric stereo by spectral multiplexing.
\newblock In {\em Proc.~ICCP}, 2011.

\bibitem{Hayakawa94:PS}
H.~Hayakawa.
\newblock Photometric stereo under a light source with arbitrary motion.
\newblock {\em JOSA A}, 1994.

\bibitem{Horn70:SFS}
B.~K.~P. Horn.
\newblock {\em Shape from shading; a method for obtaining the shape of a smooth
  opaque object from one view}.
\newblock PhD thesis, Massachusetts Institute of Technology, 1970.

\bibitem{Zsolt10:Dynamic}
Z.~Janko, A.~Delaunoy, and E.~Prados.
\newblock Colour dynamic photometric stereo for textured surfaces.
\newblock In {\em Proc.~ACCV}, 2010.

\bibitem{Johnson09:SFS}
M.~K. Johnson and E.~H. Adelson.
\newblock Retrographic sensing for the measurement of surface texture and
  shape.
\newblock In {\em Proc.~CVPR}, 2009.

\newpage

\bibitem{Johnson11:SFS}
M.~K. Johnson and E.~H. Adelson.
\newblock Shape estimation in natural illumination.
\newblock In {\em Proc.~CVPR}, 2011.

\bibitem{Kontsevich94:SFS}
L.~L. Kontsevich, A.~P. Petrov, and I.~S. Vergelskaya.
\newblock Reconstruction of shape from shading in color images.
\newblock {\em JOSA A}, 1994.

\bibitem{Vlasic09:DynamicShape}
D.~Vlasic, P.~Peers, I.~Baran, P.~Debevec, J.~Popovi\'{c}, S.~Rusinkiewicz, and
  W.~Matusik.
\newblock Dynamic shape capture using multi-view photometric stereo.
\newblock {\em ACM Trans. Graph. (SIGGRAPH Asia)}, 2009.

\bibitem{Woodham80:PS}
R.~Woodham.
\newblock Photometric method for determining surface orientation from multiple
  images.
\newblock {\em Optical Engineering}, 1980.

\bibitem{Woodham94:RGBPS}
R.~J. Woodham.
\newblock Gradient and curvature from the photometric-stereo method, including
  local confidence estimation.
\newblock {\em JOSA A}, 1994.

\bibitem{qsfs}
Y.~Xiong, A.~Chakrabarti, R.~Basri, S.~J. Gortler, D.~W. Jacobs, and
  T.~Zickler.
\newblock From shading to local shape.
\newblock {\em IEEE Trans. PAMI}, 2015.

\bibitem{Zhang99:SFSSurvey}
R.~Zhang, P.-S. Tsai, J.~E. Cryer, and M.~Shah.
\newblock Shape from shading: A survey.
\newblock {\em IEEE Trans. PAMI}, 1999.

\end{thebibliography}

}

\end{document}